\documentclass[sigconf,nonacm]{acmart}
\renewcommand\footnotetextcopyrightpermission[1]{} 
\usepackage{url}    
\usepackage{listings}
\usepackage{xcolor}
\usepackage{makecell} 
\usepackage{float} 
\usepackage{stfloats} 
\usepackage{placeins} 
\usepackage{multirow} 
\usepackage{enumitem}
\usepackage{array}  
\usepackage{booktabs}
\usepackage{tabularx}
\usepackage{booktabs}
\usepackage{siunitx}
\usepackage{mathtools} %
\usepackage[font=small,skip=2pt]{caption} 
\AtBeginDocument{%
  }

\setcopyright{acmlicensed}
\copyrightyear{2018}
\acmYear{2018}
\acmDOI{XXXXXXX.XXXXXXX}

\acmConference[Conference acronym 'XX]{Make sure to enter the correct
  conference title from your rights confirmation emai}{June 03--05,
  2018}{Woodstock, NY}
\acmISBN{978-1-4503-XXXX-X/18/06}

\begin{document}

\title{HADSF: Aspect Aware Semantic Control for Explainable Recommendation}


\author{Zheng Nie}
\affiliation{%
  \institution{National University of Singapore}
  \city{Singapore}
  \country{Singapore}}
\email{zhengnie02@gmail.com}

\author{$^{\dagger}$Peijie Sun}
\affiliation{%
  \institution{Nanjing University of Posts and Telecommunications}
  \city{Nanjing}
  \country{China}}
\email{sun.hfut@gmail.com}

\thanks{$^{\dagger}$ Corresponding author.}

\renewcommand{\shortauthors}{Trovato et al.}

\begin{abstract}
Recent advances in large language models (LLMs) promise more effective information extraction for review-based recommender systems, yet current methods still (i) mine free-form reviews without scope control, producing redundant and noisy representations, (ii) lack principled metrics that link LLM hallucination to downstream effectiveness, and (iii) leave the cost–quality trade-off across model scales largely unexplored. We address these gaps with the Hyper-Adaptive Dual-Stage Semantic Framework (HADSF), a two-stage approach that first induces a compact, corpus-level aspect vocabulary via adaptive selection and then performs vocabulary-guided, explicitly constrained extraction of structured aspect-opinion triples. To assess the fidelity of the resulting representations, we introduce Aspect Drift Rate (ADR) and Opinion Fidelity Rate (OFR) and empirically uncover a nonmonotonic relationship between hallucination severity and rating prediction error. Experiments on approximately 3 million reviews across LLMs spanning 1.5B-70B parameters show that, when integrated into standard rating predictors, HADSF yields consistent reductions in prediction error and enables smaller models to achieve competitive performance in representative deployment scenarios. We release code, data pipelines, and metric implementations to support reproducible research on hallucination-aware, LLM-enhanced explainable recommendation. Code is available at https://github.com/niez233/HADSF.
\end{abstract}


\keywords{Explainable Recommendation; Large Language Models; Semantic Aspect;  Hallucination.}

\maketitle

\section{Introduction}
Recommender systems have become indispensable infrastructure in the digital economy, with global e-commerce platforms processing over 12 billion product recommendations daily and contributing to approximately 35\% of Amazon's revenue and 80\% of Netflix's viewing time~\cite{chen2024reasoner,li2023personalized}. As these systems increasingly influence user decision-making across domains from entertainment to healthcare, the demand for explainable recommendations has intensified. Modern users not only expect accurate suggestions but also require transparent justifications that foster trust and enable informed choices~\cite{Shuai2023TopicenhancedGN,zhang2014explicit}. This paradigm shift has positioned explainable recommender systems as a critical research frontier, where the dual objectives of prediction accuracy and interpretability must be optimized simultaneously.

\begin{figure}[htbp]
    \centering
    \includegraphics[width=0.9\linewidth]{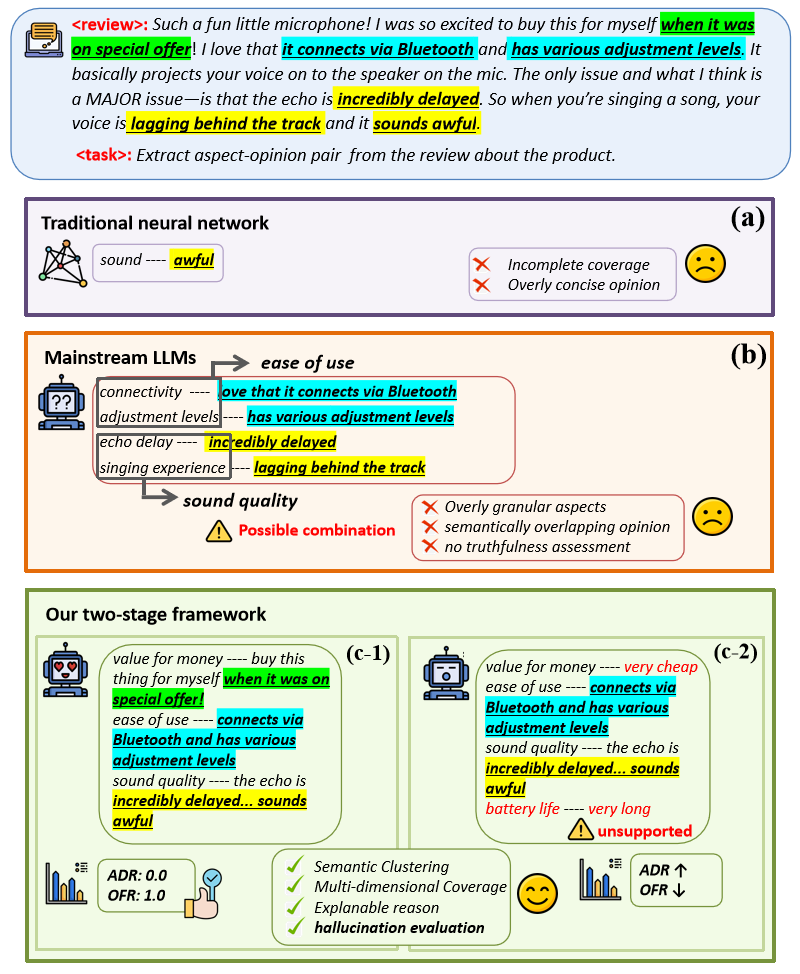}
    \caption{Comparison of aspect–opinion extraction across traditional neural models, mainstream LLM methods and our HADSF. Color‐coded highlights denote different aspect–opinions.}
    \label{fig:intro}
\end{figure}

Building on this demand for interpretability, a major line of work has focused on harnessing user-generated reviews to capture preference signals and provide explanatory content, with approaches evolving from traditional neural architectures to recent large language models (LLMs). Traditional approaches to explainable recommendation have predominantly leveraged user-generated reviews as rich sources of preference signals and explanatory content~\cite{xi2021deep,geng2022recommendation}. These review-based methods generally fall into two categories: general review-based approaches that extract holistic user preferences~\cite{Zheng2017JointDM,Chen2018NeuralAR,Liu2019DAMLDA}, and aspect-based methods that decompose reviews into fine-grained opinion dimensions~\cite{zhang2022survey,li2020towards}. While early neural architectures achieved reasonable performance through topic modeling and attention mechanisms~\cite{cheng2018aspect, 10.1145/3269206.3271810, Guan2018AttentiveAM}, they often struggled with semantic understanding and generated explanations that lacked coherence or relevance to actual user reasoning processes. As illustrated in Fig.~\ref{fig:intro}(a), such approaches tend to produce fragmented or semantically inconsistent aspect–opinion pairs, highlighting the inherent limitations in capturing true user intent. The emergence of large language models (LLMs) has catalyzed a paradigm shift in review-based recommendation, offering unprecedented capabilities in text understanding and generation~\cite{Ren2023RepresentationLW,tsai2024leveragingllmreasoningenhances,bismay2024reasoningrec}. Recent work explores transforming raw reviews into structured preference representations~\cite{kim2024driven,fang2025reason4reclargelanguagemodels} and employing post-training techniques for aspect extraction~\cite{li2023prompttuninglargelanguage}. These LLM-based approaches generally surpass traditional neural methods in semantic alignment and interpretability. 

Nevertheless, current LLM-enhanced recommenders face three core limitations that hinder real-world applicability and scalability. First, existing methods lack principled control over the scope and granularity of information extraction, often decomposing reviews into overly fragmented, attribute-level tags as illustrated in Fig.\ref{fig:intro}(b). This uncontrolled granularity produces semantically overlapping tags for the same underlying issue, inflating redundancy, and also yields extremely low-frequency attributes that act as statistical noise in downstream recommenders—effects that are especially acute on large platforms where reviews range from terse one-liners to multi-paragraph narratives. Compounding this, current pipelines~\cite{liu2025understanding} rarely verify the truthfulness of extracted content or leverage contextual priors at the user or item level that could guide more reliable extraction. Second, existing pipelines provide no rigorous, quantitative assessment of the factual fidelity of extracted content. Current hallucination evaluation practices, such as counting out-of-catalog items in generated lists~\cite{jiang2025beyond} fail to capture how hallucinations propagate into downstream recommendation errors. This gap precludes systematic diagnosis of when an LLM’s fabricated or drifted content materially degrades predictive performance, and thus prevents principled mitigation strategies. Third, while state-of-the-art, large-scale LLMs can achieve high extraction accuracy, their prohibitive inference costs limit deployment in latency- or budget-sensitive scenarios. Conversely, smaller, cost-effective models tend to exhibit higher hallucination rates, yet the cost–quality trade-off—how model scale jointly influences hallucination severity and recommendation quality—remains unexplored. Bridging these gaps requires not only controlled extraction to reduce redundancy and noise, but also interpretable hallucination metrics that can be directly correlated with downstream performance, thereby enabling informed model selection under real-world cost constraints.

To address these challenges, we propose the Hyper-Adaptive Dual-Stage Semantic Framework (HADSF) which transforms free-form reviews into controlled, structured aspect-opinion representations while systematically measuring and analyzing LLM-induced hallucination effects. Stage I induces a compact, corpus-level aspect vocabulary via adaptive selection over user review corpora. Stage II uses this vocabulary as explicit constraints to extract structured aspect–opinion triples, incorporating user and item contextual priors. Crucially, we introduce two interpretable hallucination metrics—Aspect Drift Rate (ADR) and Opinion Fidelity Rate (OFR)—to quantify how severely the LLM fabricates content during extraction, and we correlate these rates with downstream rating prediction error to reveal a non-monotonic hallucination–performance relationship. As shown in Fig.\ref{fig:intro}(c-1) and (c-2), HADSF yields more coherent and non-redundant aspect–opinion triples than traditional neural and mainstream LLM baselines. Fig.\ref{fig:intro}(c-1) shows the ideal results of structured extraction conditioned on a corpus-level aspect vocabulary and contextual priors, while Fig.\ref{fig:intro}(c-2) overlays the progress of hallucination diagnostics. Finally, we benchmark models from 1.5B to 70B parameters and show that smaller LLMs can match or even exceed larger counterparts in several regimes—providing actionable guidance for cost-constrained, large-scale deployment under HADSF.

Our primary contributions are threefold: (1) We develop the first systematic framework for controlled aspect-opinion extraction in large-scale recommendation, demonstrating consistent improvements in prediction accuracy across diverse review characteristics and dataset scales; (2) We introduce novel, interpretable metrics for quantifying LLM hallucination in recommendation contexts and provide the first empirical analysis revealing a non-monotonic relationship between hallucination severity and recommendation performance; (3) We conduct comprehensive experiments on LLMs ranging from 1.5B to 70B parameters over datasets totaling approximately 3 million reviews, uncovering scale-dependent hallucination patterns, and showing the viability of small language models under HADSF's extraction. We
release an open-source toolkit with code, data pipelines and metric implementations to support reproducible research and establish new benchmarks for LLM-enhanced explainable recommendation.

The remainder of this paper is organized as follows: Section 2 reviews related work, Section 3 presents our methodology, Section 4 details experimental setup and results, and Section 5 concludes with implications and future directions.

\section{Related Work}
\label{sec:relatedwork}
In this section, we conduct a comprehensive literature review focused on LLMs in recommendation and review-based recommendation approaches. Our research is inspired by these themes, exploring the integration of LLMs with review-based recommendation strategies to enhance the personalization and accuracy of suggested content.

\subsection{Review-based Recommendation}
Previous work on review-based recommendation aims to address data scarcity and enhance explainability~\cite{Chen2023DataAS,wei2024llmrec} by extracting ground-truth explanations from user–item reviews as auxiliary information. Reviews contain rich semantic details related to user ratings, enabling their analysis to refine preference models and improve prediction accuracy~\cite{Sun2020DualLF,Pan2022AccurateAE,Pugoy2024NEARNE}. Early studies often employed topic models to derive latent feature distributions for users and items, such as CTR~\cite{Wang2011CollaborativeTM}, which learns interpretable latent structures by combining topic factors with free embeddings, and TIM~\cite{Pea2020CombiningRA}, which builds embeddings within the topic space learned from review data. With the advent of deep learning, more advanced text encoders have been adopted to extract semantic information from reviews, including the collaborative deep learning framework of Wang et al.\cite{Wang2014CollaborativeDL} integrating Stacked Denoising AutoEncoders with probabilistic matrix factorization, DeepCoNN\cite{Zheng2017JointDM} employing two parallel TextCNNs~\cite{Kim2014ConvolutionalNN} to capture semantic features, NARRE~\cite{Chen2018NeuralAR} leveraging attention to score each review, and DAML~\cite{Liu2019DAMLDA} applying dual attention mutual learning to fuse rating and review features for enhanced predictions.

\subsection{LLMs in Recommendation}

Large language models (LLMs) are increasingly used in recommendation for their strong ability to model semantics and user intent. Recent LLM-based recommendation systems can be classified into two categories: discriminative and generative ~\cite{Wu2023ASO,Fan2023RecommenderSI,Bao2023LargeLM}. 
Discriminative approaches are primarily utilized to obtain refined representations of users and items by fine-tuning and exploring training strategies like prompt tuning and adapter tuning~\cite{Xiao2021TrainingLN,Hou2022TowardsUS}. Generative approaches leverage LLMs’ text generation to directly produce recommendation outputs~\cite{geng2022recommendation,Ren2023RepresentationLW,Dai2023UncoveringCC}. Building on the exceptional natural language processing capabilities, LLMs has achieved many breakthroughs in recommendation scenarios. CLLM4Rec~\cite{Zhu2023CollaborativeLL} is the first generative recommendation system(RS) that closely combines the LLM paradigm and the ID paradigm of RS, effectively alleviating the problem of invalid language modeling. P5~\cite{geng2022recommendation} converts interaction data into textual prompts using item indexes. LLaRA~\cite{Liao2023LLaRALL} fuses ID-based item embeddings with textual features to bridge conventional recommenders and LLMs. 
Despite these advances, much work still relies on ID-based embeddings or interaction logs and underutilizes the rich information in review texts.
\section{Methodology}
 
\begin{figure*}[t]
  \centering
  \includegraphics[width=\textwidth]{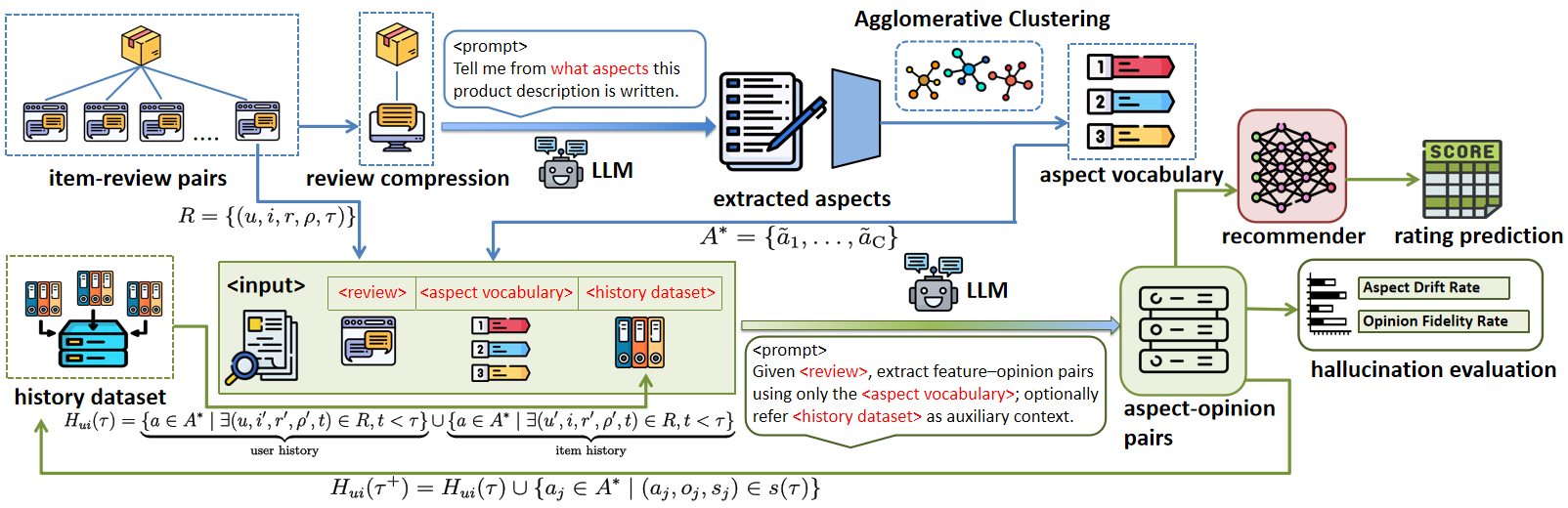}
  \caption{Framework overview. In the above section, reviews are compressed, global aspects are extracted, and the Top-N personalized aspects are selected; In the below section, the selected aspects and the target review are fed into an LLM to extract fine-grained aspect–opinion pairs and dynamically update the user’s aspect memory.}
  \label{fig:framework}
\end{figure*}

This section presents our Hyper-Adaptive Dual-Stage Semantic Framework (HADSF) for LLM-enhanced explainable recommendation with systematic hallucination quantification. Our approach addresses the fundamental challenge of extracting reliable, personalized aspect-opinion insights from user reviews while maintaining controllability over LLM-generated content. We begin with formal problem formulation, followed by detailed exposition of our framework and novel hallucination quantification metrics.

\subsection{Theoretical analysis and Framework Overview}

\subsubsection{Theoretical analysis.}
Let $\mathbf z_{\mathrm{free}}=\Phi\!\circ E_{\mathrm{free}}(X)$ and $\mathbf z_{\mathrm{ctrl}}=\Phi\!\circ E_{\mathrm{ctrl}}^{A^{*}}(X;H_{ui}(\tau))$, where $A^{*}$ is the controlled aspect codebook and $K\triangleq|A^{*}|$.
For capacity-limited decoders $\mathcal G$, we focus on the \emph{attainable} mutual information
\[ \max_{E,\Phi,\,g\in\mathcal G}\ \underbrace{I_{\mathcal G}(\mathbf z;Y)}_{\text{usable relevance}} \;-\;\beta\,\underbrace{I(\mathbf z;X)}_{\text{compression}}. \tag{1} \]
Semantic control (projection to $A^{*}$) yields
\[
I(\mathbf z_{\mathrm{ctrl}};X)\ \le\ I(\mathbf z_{\mathrm{free}};X),\
\dim(\mathbf z_{\mathrm{ctrl}})=K\ \ll\ \dim(\mathbf z_{\mathrm{free}}).
\tag{2}
\]
i.e., lower entropy and lower dimension. With $n$ samples and a fixed decoder family, the test risk obeys the standard form
\[
\mathcal{E}_{\mathrm{test}}(g)\ \lesssim\ \mathcal{E}_{\mathrm{train}}(g)\;+\;\mathcal{O}\!\Big(\sqrt{\tfrac{\mathrm{Comp}(\mathbf z)}{n}}\Big).
\tag{3}
\]
where $\mathrm{Comp}(\mathbf z)$ is a capacity/complexity term that increases with $I(\mathbf z;X)$ and with $\dim(\mathbf z)$.
Combining (1)–(3), the controlled representation $\mathbf z_{\mathrm{ctrl}}$ \emph{tightens} the generalization term by reducing $I(\mathbf z;X)$ and dimensionality, and—so long as the task-relevant information $I_{\mathcal G}(\mathbf z;Y)$ is preserved or improved by aligning to $A^{*}$—it yields a lower downstream test error.
This establishes the necessity of introducing a compact, task-aligned aspect space before recommender learning.

\subsubsection{Framework Architecture}  As illustrated in Fig.~\ref{fig:framework}, our HADSF operates through two synergistic stages: (1) Controlled Semantic Aspect Extraction that establishes a domain-specific aspect vocabulary through multi-sampling consensus, and (2) Dynamic Aspect-Aware Review Processing that leverages personalized historical context for targeted extraction. This design ensures both global consistency and individual adaptability, addressing the inherent tension between standardization and personalization in recommendation systems.

\subsection{Stage I: Controlled Semantic Aspect Extraction}

\subsubsection{Multi-Sampling Consensus Mechanism}
To establish a robust and representative aspect vocabulary while mitigating LLM hallucination, we introduce a multi-sampling consensus approach. Let the complete review corpus be
$D = \{r_1, r_2, \ldots, r_{|D|}\}.$
We perform \(K\) independent random subsampling steps:
\[
D^{(k)} = \operatorname{Subsample}\bigl(D,\tfrac{|D|}{K}\bigr),
\substack{\quad D^{(1)},\dots,D^{(K)}\ \text{i.i.d.},\quad k=1,\dots,K.}
\tag{4}
\]
Each sampled subset is preprocessed to obtain cleaned review sequences and then prompt an LLM to compress the original long text into a high information–density abstract:
\[
\tilde T_i^{(k)}
= L_M\!\Bigl(P_{\mathrm{abs}}\bigl(\big\Vert_{r\in D_i^{(k)}} r\bigr)\Bigr),\quad
\substack{i=1,\dots,N\\k=1,\dots,K}.
\tag{5}
\]
Finally, aspects are extracted from each sample-specific abstract:
\begin{align}
A^{(k)} &= L_M\!\Bigl(P_{\mathrm{aspect}}\bigl[\tilde T^{(k)}\bigr]\Bigr)\nonumber\\
        &= \{a_1, \ldots, a_{M_k}\},\qquad k = 1,\ldots,K.\tag{6}
\end{align}

\subsubsection{Semantic Normalization via Embedding-Based Clustering}
Raw aspect extraction often yields semantically similar terms (e.g., ``ambiance,'' ``atmosphere,'' ``environment'') that represent the same concept. 
To consolidate such near-duplicates, we perform clustering-based normalization.

Let $\phi(\cdot)$ be a pre-trained sentence encoder. We first embed each distinct aspect term $a'$ from $\bigcup_{k=1}^K A^{(k)}$ as $\mathbf{z}_{a'} = \phi(a')$, and then apply agglomerative clustering on $\{\mathbf{z}_{a'}\}_{a'}$ to group semantically similar aspects into $C$ clusters:
\begin{align}
\{C_1, \ldots, C_C\} &= \text{AgglomerativeClustering}\big(\{\mathbf{z}_{a'}\}_{a'}, C\big). \tag{7} \label{eq:cluster}
\end{align}

Within each cluster $C_c$ ($c=1,\ldots,C$), let $\mathrm{freq}(a)$ be the occurrence count of aspect $a$.
We select the representative aspect $\tilde a_c$ by maximizing the frequency-normalized semantic similarity to all other aspects in the same cluster:
\begin{align}
\tilde a_c
= \arg\max_{a' \in C_c}
\sum_{a'' \in C_c} 
\frac{\mathrm{freq}(a'')}{\sum_{x \in C_c} \mathrm{freq}(x)}\, \hat\phi(a')^\top \hat\phi(a'').
\tag{8}\label{eq:centroid}
\end{align}

\begin{figure}[t]
    \centering
    \begin{minipage}{0.49\linewidth}
        \centering
        \includegraphics[width=\linewidth]{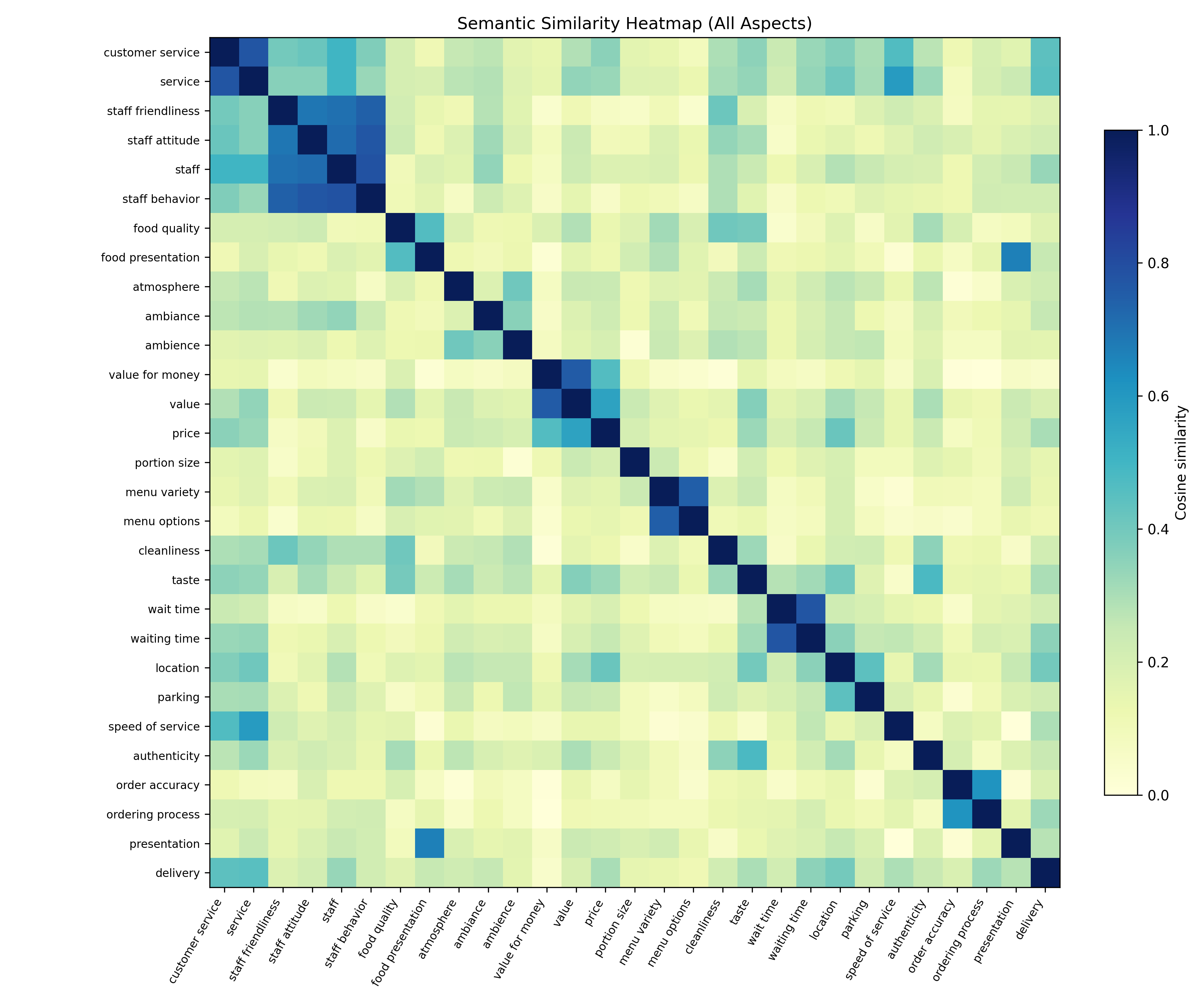}
        \textbf{(a)} Semantic similarity heatmap \textbf{before} aspect clustering.
        \label{fig:before_cluster}
    \end{minipage}
    \hfill
    \begin{minipage}{0.49\linewidth}
        \centering
        \includegraphics[width=\linewidth]{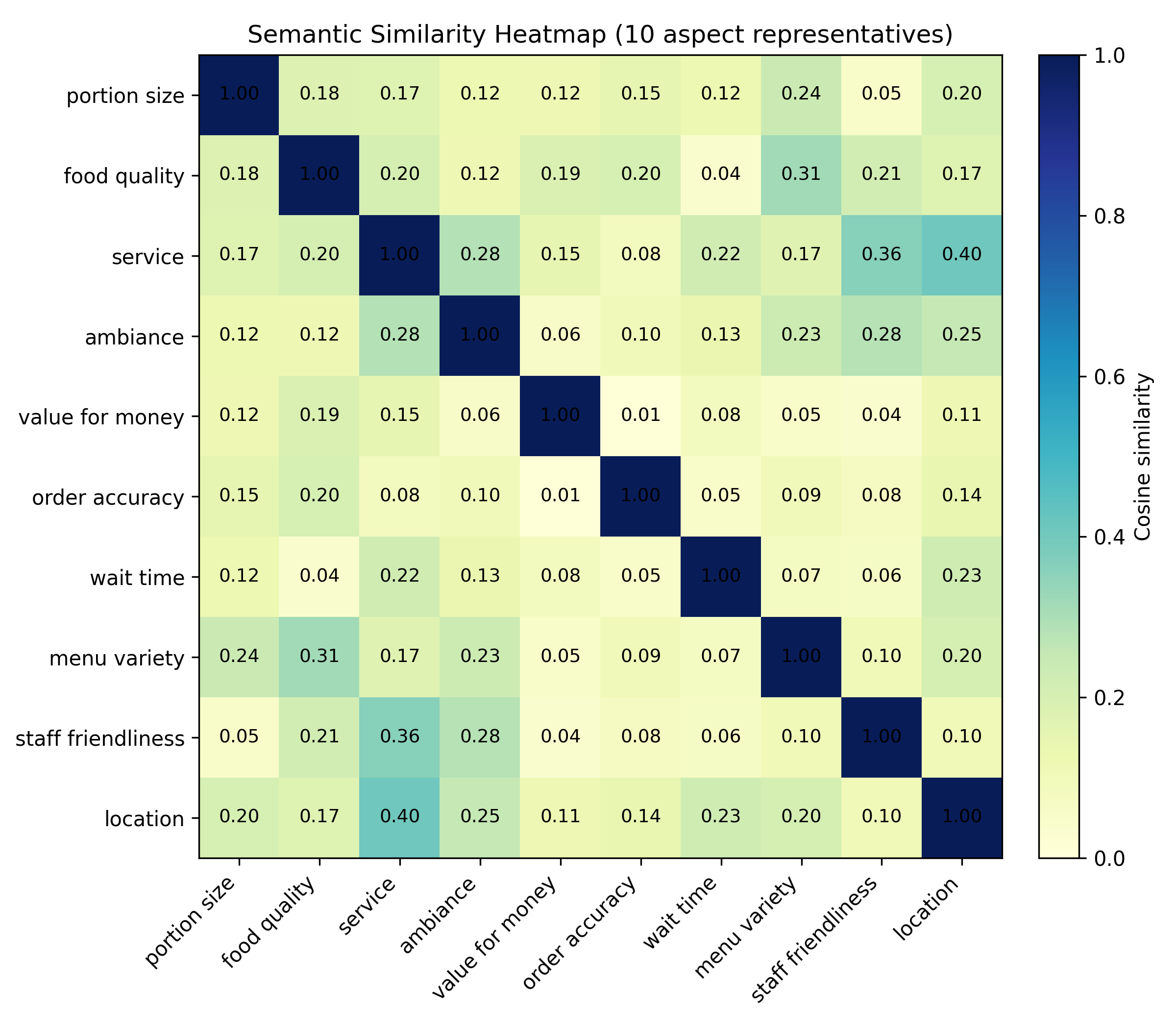}
        \textbf{(b)} Semantic similarity heatmap \textbf{after} aspect clustering.
        \label{fig:after_cluster}
    \end{minipage}
    \vspace{0.5em}    
    \caption{Comparison of feature independence of semantic aspects before and after clustering.}
    \label{fig:aspect_cluster_analysis}
\end{figure}

Where $\hat\phi(x)=\phi(x)/\|\phi(x)\|_2$ and $a',a'' \in C_c$ index aspects within the same cluster. The final consolidated vocabulary is:
\begin{align}
A^{*} = \{\tilde a_1, \ldots, \tilde a_C\}. \tag{9} \label{eq:finalA}
\end{align}
As illustrated in Fig.~\ref{fig:aspect_cluster_analysis}, this clustering-based normalization reduces redundancy by grouping semantically similar aspects, leading to more diverse and representative aspect sets.

\subsection{Stage II: Dynamic Aspect-Aware Review Processing}

\subsubsection{Temporal Personalization Strategy}
Building upon the distilled aspect vocabulary $A^{*}$, our second stage implements dynamic, personalized extraction that evolves with both user preferences and item characteristics over time. This addresses a critical limitation in existing approaches: the failure to account for temporal dynamics in user–item interactions.

We sort all interactions $R$ in chronological order by timestamp $\tau$ and maintain a dynamic history dataset
$H_{ui} = \{\}$,
where $H_{ui}\subseteq A^{*}$ accumulates aspects previously attended to by user $u$ and attributed to item $i$.

For each interaction $(u, i, r, \rho, \tau)\in R$ processed chronologically, we construct a personalized prompt that integrates global aspect guidance with the combined user–item historical context:
\begin{align}
P_{\text{dynamic}}
&= \bigl[\,P_{\text{global}}[A^{*}]\;;\;P_{\text{personal}}[H_{ui}(\tau)]\;;\;P_{\text{extract}}[r]\,\bigr], 
\tag{10}\label{eq:prompt}
\end{align}
where
\[
\begin{aligned}
H_{ui}(\tau)
&= \bigl\{\,a\in A^* \mid 
\exists\,(u,i',r',\rho',t)\in R,\;t<\tau
\bigr\} \\[4pt]
&\quad\;\cup\;
\bigl\{\,a\in A^* \mid 
\exists\,(u',i,r',\rho',t)\in R,\;t<\tau
\bigr\}.
\end{aligned}\tag{11}
\]
represents the user’s historical aspect focus derived from their past interactions and the item’s characteristic aspects emphasized by other users.

\subsubsection{Controlled Extraction Process}
The LLM processes the dynamic prompt to generate structured aspect-opinion-sentiment triples:
\begin{align}
s &= L_M\!\big(P_{\text{dynamic}}\big) = \{(a_j, o_j, s_j) \mid j = 1, \ldots, |s|\}. \tag{12} \label{eq:triples}
\end{align}
where each triple $(a_j, o_j, s_j)$ represents an extracted aspect, corresponding opinion, and sentiment polarity, respectively.

To maintain temporal consistency and enable progressive personalization, we update the history dataset:
\begin{align}
H_{ui}(\tau^{+})
&= H_{ui}(\tau)\;\cup\;
\bigl\{\, a_j \in A^{*} \;\bigm|\; (a_j, o_j, s_j) \in s(\tau) \,\bigr\}.\tag{13} \label{eq:history}
\end{align}

This dynamic updating mechanism ensures that subsequent extractions for the same user benefit from accumulated preference knowledge while remaining adaptable to evolving interests.

\subsection{Hallucination Quantification Framework}

\subsubsection{Novel Hallucination Metrics}
A critical innovation of our framework lies in systematic quantification of LLM hallucination effects. We introduce two complementary metrics:

\textbf{Aspect Drift Rate (ADR):} Measures the proportion of extracted aspects that deviate from the established vocabulary $A^{*}$:
\begin{align}
\text{ADR} &= \frac{1}{|S|} \sum_{(u,i,\tau,s) \in S} \frac{\big|\{a \mid (a,o,s) \in s,\; a \notin A^{*}\}\big|}{|s|}. \tag{14} \label{eq:adr}
\end{align}

\textbf{Opinion Fidelity Rate (OFR):} Evaluates the semantic consistency between extracted opinions and original review content:
\begin{align}
\text{OFR} &= \frac{1}{|S|} \sum_{(u,i,\tau,s) \in S} \frac{1}{|s|} \sum_{(a,o,s) \in s} \text{SemSim}(o, r). \tag{15} \label{eq:ofr}
\end{align}
Here, $\text{SemSim}$ implements span-level grounding by identifying, for each extracted opinion $o$, the contiguous span in the source review $r$ that achieves the highest semantic similarity. We define $\text{SemSim}(o,r)$ as
\begin{align}
\text{SemSim}(o,r) &=
\max_{r_{s:e} \in \mathcal{S}_{L(o),\Delta}(r)}
\frac{\langle \psi(o),\, \psi(r_{s:e}) \rangle}{
\lVert \psi(o) \rVert_2 \, \lVert \psi(r_{s:e}) \rVert_2 }.
\tag{16}\label{eq:semsim_span}
\end{align}

where $\mathcal{S}_{L,\Delta}(r)$ enumerates all contiguous spans $r_{s:e}$ in $r$ whose token length lies within $[L(o)-\Delta,\, L(o)+\Delta]$. We write $r = x_{1:m}$ for a review of $m$ tokens, and $L(o)$ denotes the number of tokens in $o$. The function $\psi(x_{1:T})$ computes the mean of token embeddings:
\begin{align}
\psi(x_{1:T}) &\triangleq \frac{1}{T} \sum_{t=1}^T \phi(x_t)\tag{17}
\end{align}
where $\phi(\cdot)$ denotes the embedding function. The bracket $\langle \cdot, \cdot \rangle$ denotes the Euclidean inner product; with the normalization in \eqref{eq:semsim_span}, $\mathrm{SemSim}(o,r)$ is the \emph{cosine similarity} between mean embeddings, capturing directional alignment while being invariant to vector norms. If the opinion text $o$ appears verbatim in $r$ (case-insensitive), we set $\mathrm{SemSim}(o,r)=1$.

\begin{figure}[t]
    \centering
    \includegraphics[width=\linewidth]{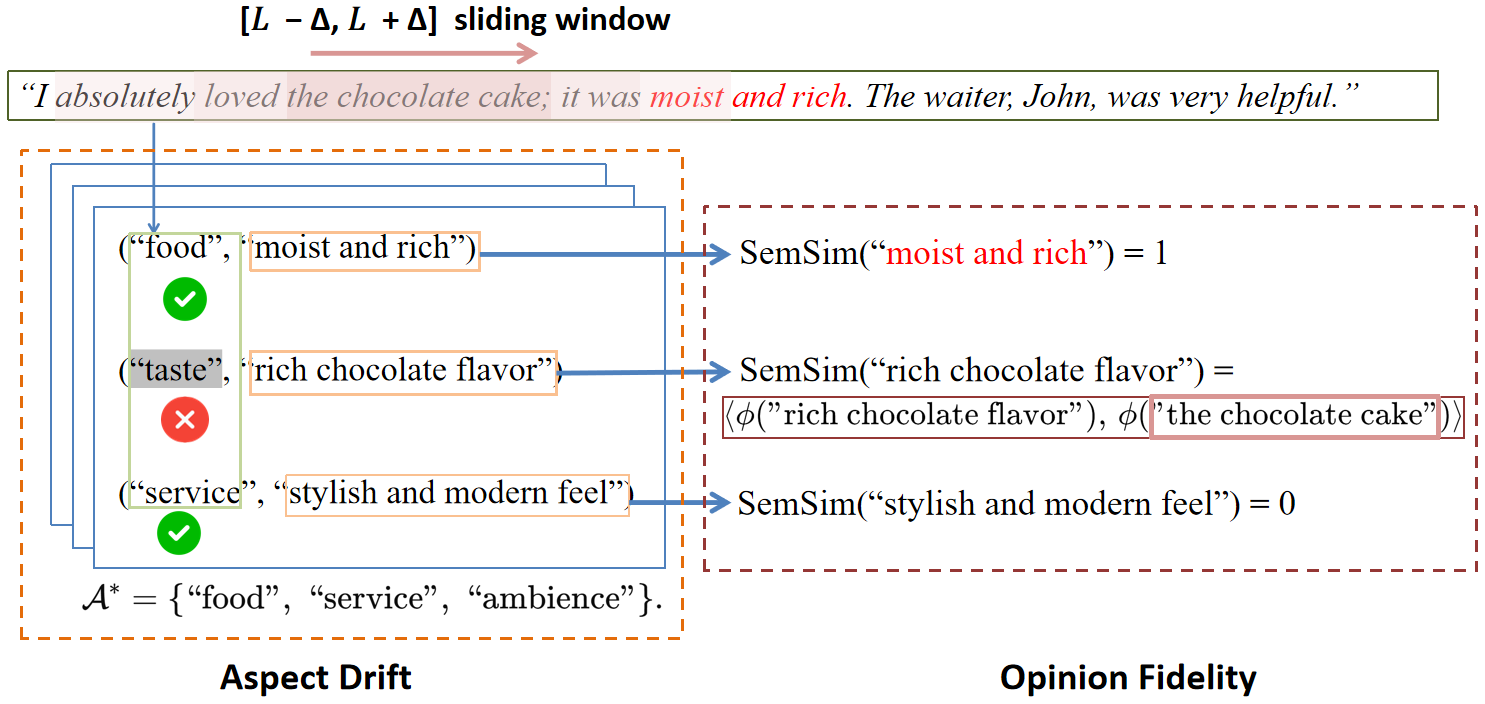}
    \caption{Identification process of Aspect Drift and Opinion Fidelity. }
    \label{fig:adr}
\end{figure}

\section{Experiments}
\label{sec:experiments}
In this section, we conduct a series of experiments to answer the
following research questions:
\begin{itemize}[leftmargin=*, align=left]
\item \textbf{RQ1:} How does our review extraction framework compare to baseline methods in rating prediction?
\item \textbf{RQ2:} How do different review extraction strategies impact recommendation performance?
\item \textbf{RQ3:} How does the degree of hallucination in the extracted reviews affect downstream recommendation outcomes?
\end{itemize}

\subsection{Experimental Settings}
\subsubsection{Datasets}We conduct experiments on three datasets, including two datasets from Amazon\footnote{\url{http://jmcauley.ucsd.edu/data/amazon}} and another one from Yelp\footnote{\url{https://www.yelp.com/dataset}}. All three datasets contain user interaction records with items, including user IDs, item IDs, textual reviews and ratings. To control dataset size, we apply the following filters: for Amazon, we retain users and items with more than 25 records since 2021; for Yelp, we focus on restaurant-category records from 2019 onward and keep users and items with more than 10 records. The statistics of these three datasets are summarized in \hyperref[tab:stats]{Table~\ref*{tab:stats}}.

\subsubsection{Evaluation Metric}We evaluate our method on rating prediction models using Mean Squared Error (MSE) and Mean Absolute Error (MAE), which are widely adopted metrics to quantify the deviation between predicted and true rating values.

\subsubsection{Baselines}
For rating prediction, we select a set of both tradi-
tionally used and recently proposed review-based baselines, which can briefly categorized into three groups: (1) \textit{Traditional rating-based collaborative filtering methods}: PMF~\cite{mnih2007probabilistic} and EFM~\cite{zhang2014explicit}; (2) \textit{Review-aware neural network methods}: ANR~\cite{10.1145/3269206.3271810}, NARRE~\cite{Chen2018NeuralAR}, DeepCoNN~\cite{Zheng2017JointDM}, RGCL~\cite{shuai2022review} and TGNN~\cite{Shuai2023TopicenhancedGN}; (3) \textit{LLM-as-Recommender methods}: GPT-4o~\cite{liu2023chatgpt}, Rec-SAVER~\cite{tsai2024leveragingllmreasoningenhances} and EXP3RT~\cite{kim2024driven}. For detailed descriptions of baseline and implementation details, please refer to Appendix A.

\subsubsection{Aspect-aware integration} To assess generality and effectiveness, we integrate our aspect-aware review extraction module (Section 3) into EFM, ANR, and TGNN by feeding the extracted aspects into their input pipelines, and we evaluate whether this integration consistently improves recommendation performance.

\begin{table}[t] 
\centering 
\caption{Statistics of datasets.} 
\setlength{\aboverulesep}{0.3ex}   
\setlength{\belowrulesep}{0.25ex}   
\renewcommand{\arraystretch}{0.98} 
\label{tab:stats} 
\begin{tabularx}{\columnwidth}{>{\raggedright\arraybackslash}X *{3}{>{\centering\arraybackslash}X}} 
\toprule[1pt]
Statistic & Musical & Industrial & Yelp \\
\midrule
\#Users & 320,689 & 884,066 & 599,070 \\
\#Items & 15,273 & 25,576 & 44,904 \\
\#Reviews & 402,326 & 1,043,019 & 1,481,005 \\
\#Sentences & 1,390,847 & 3,095,899 & 10,970,203 \\
Avg Review Len. & 42.52 & 34.82 & 90.03 \\
Density & 0.183\% & 0.078\% & 0.075\% \\
\bottomrule[1pt]
\end{tabularx}
\end{table}

\subsection{Performance Comparison(RQ1)}
\subsubsection{Main Results (vs. Baselines)}
Based on the results in Table \ref{tab:llm-prediction}, our aspect‑extraction framework delivers clear, architecture‑agnostic gains in user‑feedback prediction. Across all three datasets and backbones (EFM, ANR, TGNN), the “‑aspect” variants uniformly reduce both MSE and MAE, confirming that the module is plug‑and‑play and easy to integrate. In absolute terms, TGNN‑aspect achieves the best MSE on every dataset and the second‑best MAE overall (behind RGCL), while EFM‑aspect enjoys the largest lift over its backbone. These patterns indicate that injecting structured, controllable aspects regularizes representation learning and sharpens the alignment between review content and user–item interactions, leading to accurate and stable recommendations.

Beyond conventional recommenders, the TGNN‑aspect configuration surpasses LLM‑based approaches on both error metrics across all datasets, and the remaining “‑aspect” variants remain highly competitive. By distilling reviews into concise, interpretable aspect signals rather than relying on heavy end‑to‑end generation, our method provides a modular, transferable enhancement that reliably upgrades strong backbones and yields even larger gains for weaker ones—delivering robust improvements with minimal integration effort.

\begin{table*}[!ht]
  \centering
  \caption{Comparison of user feedback prediction performance among different methods across the three evaluation datasets. The best and second-best results are highlighted in bold and underlined. * indicates performance improvements that are statistically significant with p-value < 0.05.}
  \label{tab:llm-prediction}
  \setlength{\aboverulesep}{0.3ex}   
    \setlength{\belowrulesep}{0.25ex}   
    \renewcommand{\arraystretch}{0.90} 
  \begin{tabular}{l
                  *{2}{S[table-format=1.4]}
                  *{2}{S[table-format=1.4]}
                  *{2}{S[table-format=1.4]}}
    \Xhline{1pt}
    \multirow{2}{*}{\textbf{Method}}
      & \multicolumn{2}{c}{\textbf{Musical Instruments}}
      & \multicolumn{2}{c}{\textbf{Industrial and Scientific}}
      & \multicolumn{2}{c}{\textbf{Yelp}} \\
    \cmidrule(lr){2-3} \cmidrule(lr){4-5} \cmidrule(lr){6-7}
      & {MSE} & {MAE} & {MSE} & {MAE} & {MSE} & {MAE} \\
    \midrule
    PMF           & 2.3767 & 1.2675 & 2.6328 & 1.3831 & 2.6720 & 1.4176 \\
    EFM           & 1.4842 & 0.7955 & 1.7130 & 0.8592 & 1.8686 & 0.9950 \\
    \midrule
    ANR           & 1.2960 & 0.8150 & 1.5186 & 0.8735 & 1.4108 & 0.9294 \\
    NARRE         & 1.3908 & 0.8257 & 1.6726 & 0.9379 & 1.5279 & 0.9655 \\
    DeepCoNN      & 1.3911 & 0.8065 & 1.6683 & 0.8753 & 1.6242 & 0.9742 \\
    RGCL          & 1.1557 & \textbf{0.7096} & 1.3373 & \textbf{0.7597} & 1.3537 & \textbf{0.8738} \\
    TGNN          & \underline{1.1289} & 0.7449 & \underline{1.3050} & 0.8368 & \underline{1.3470} & 0.8996 \\
    \midrule
    GPT-4o        & 1.5324 & 0.8541 & 1.6892 & 0.9023 & 1.7287 & 0.9576 \\
    Rec-SAVER     & 1.4328 & 0.8095 & 1.5785 & 0.8632 & 1.6151 & 0.9146 \\
    EXP3RT        & 1.3791 & 0.7645 & 1.5557 & 0.8373 & 1.6469 & 0.9262 \\
    \midrule
    EFM-aspect    & 1.3637* & 0.7669* & 1.5741* & 0.8342* & 1.8275* & 0.9858* \\
    ANR-aspect    & 1.2348* & 0.7953* & 1.4825* & 0.8546* & 1.3958* & 0.9271* \\
    TGNN-aspect   & \textbf{1.0960*} & \underline{0.7421*} & \textbf{1.2822*} & \underline{0.8314*} & \textbf{1.3402*} & \underline{0.8963*} \\
    \Xhline{1pt}
  \end{tabular}
\end{table*}

\subsubsection{Ablation Study}
\begin{figure}[t]
    \centering
    \includegraphics[width=0.95\linewidth]{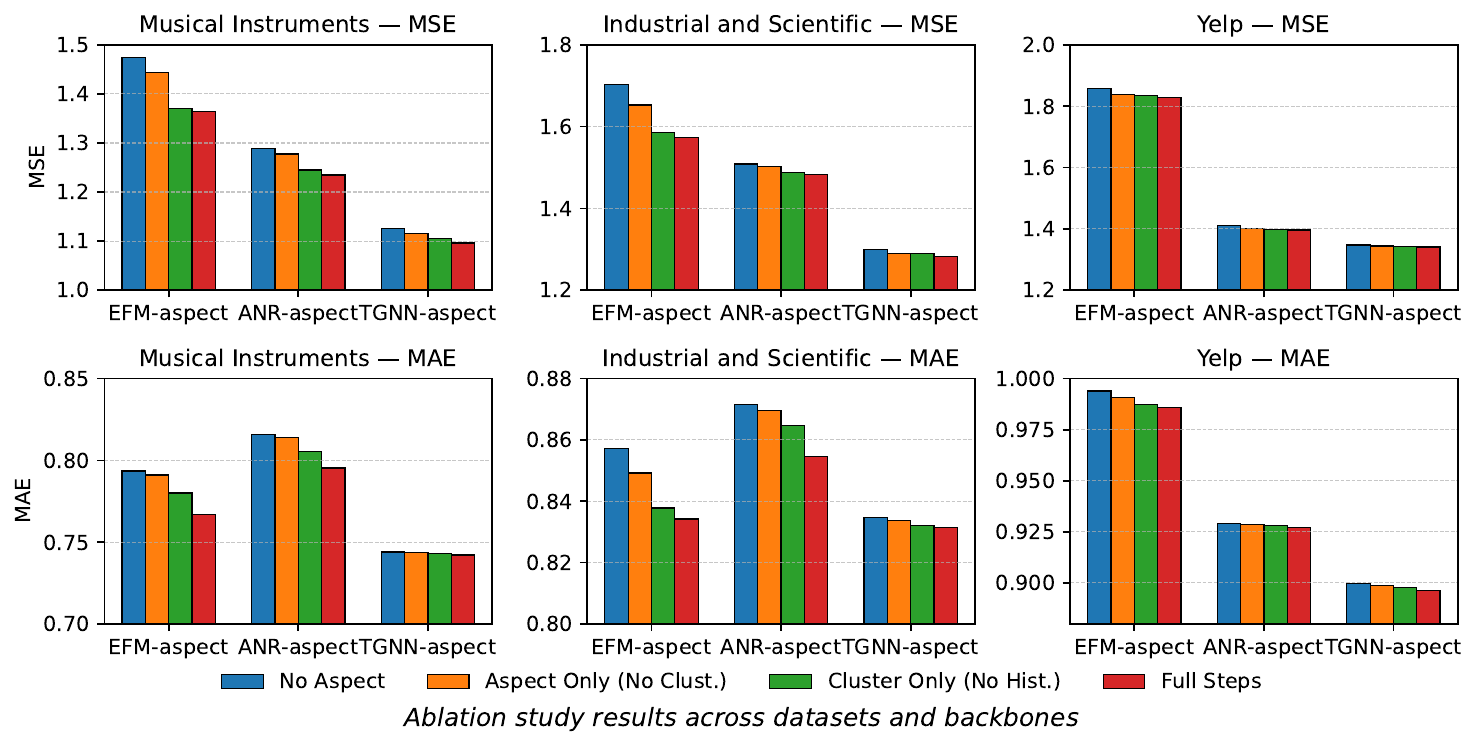}
    \caption{Ablation of the HADSF. }
    \label{fig:Ablation}
\end{figure}

Fig.~\ref{fig:Ablation} presents an ablation study of our HADSF across three backbones and datasets, evaluating the incremental impact through four variants: No Aspect (raw LLM extraction without aspect vocabulary), Aspect Only (without Clustering), Cluster Only (without Dynamic History) and Full Steps on both MSE and MAE. Results show that introducing a curated aspect vocabulary substantially reduces redundancy in raw LLM outputs, yielding marked performance gains by constraining generation to semantically relevant concepts. Further applying embedding-based clustering consolidates synonymous aspects into coherent groups, enhancing semantic normalization and improving backbone generalization across domains. These two components form the primary driver of error reduction, as they directly address the mismatch between free-form LLM generation and the structured needs of recommendation models. Building on this foundation, personalization via dynamic user–item history prompts provides consistent, complementary improvements by aligning extracted aspects with individual preference trajectories, further refining model predictions without sacrificing generalizability.


\subsection{Review Extraction Strategies (RQ2)}

\subsubsection{Aspect Quantity Configuration}
\begin{figure}[t]
    \centering
    \includegraphics[width=0.9\linewidth]{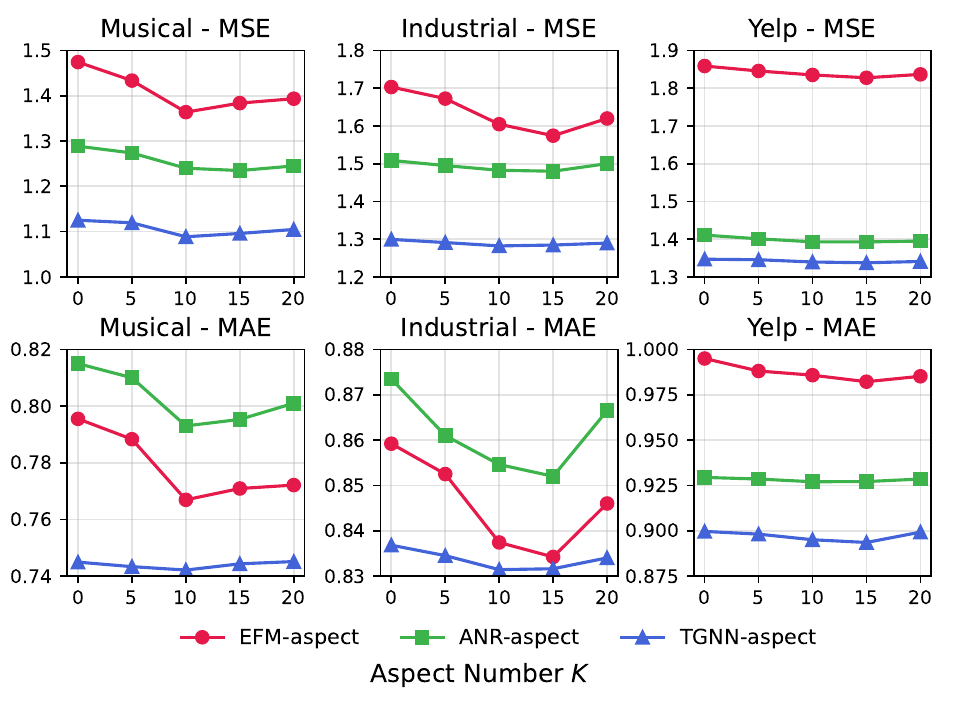}
    \caption{Performance for rating prediction under different aspect quantity configurations.}
    \label{fig:aspect_analysis}
\end{figure}

In  Fig.~\ref{fig:aspect_analysis}, we control the predefined aspect list by varying the target number of clusters $K$ during semantic aspect with $K$ set to $\{0, 5, 10, 15, 20\}$. We find that selecting $K{=}10$ or $K{=}15$ aspects yields consistently strong performance across datasets, while both smaller ($K{=}5$) and larger ($K{=}20$) configurations lead to noticeable degradation. We attribute this to the balance between coverage and precision in the aspect-aware review extraction process. With too few aspects, important semantic dimensions may be missed, limiting the diversity of extracted feature-opinion pairs. Conversely, a larger $K$ introduces infrequent or noisy aspects, increasing the risk of irrelevant extraction and diluting model focus. 

\subsubsection{Impact of Review length Categorization}
To assess the efficacy of compressed semantic representations in guiding recommendation, we stratified reviews by word count into four regimes—Short (1–10 words), Medium (10–50 words), Long (50–100 words), and Extra-long (100+ words)—and contrasted the TGNN pipeline, which constructs a sentence-vector cosine-similarity graph and applies Infomap to detect dense topic communities, against our aspect-aware framework that omits graph construction and community detection in favor of directly injecting extracted aspect–opinion pairs as compact, information-rich units. 

Table \ref{tab:metrics_no_ratio} shows the performance measured via MSE and MAE. Our aspect‐aware framework delivers the strongest predictive accuracy at both extremes of review length, outperforming TGNN on the Short and Extra-long subsets. In the Short regime, where raw text provides minimal context, our method’s focused extraction of aspect–opinion pairs succeeds in maximizing the signal‐to‐noise ratio, effectively concentrating all available information into the most discriminative features. Conversely, in the Extra‐long regime, our compression removes redundant or tangential content—common in verbose reviews and preserves only the high‐mutual‐information elements critical to rating prediction, thereby avoiding the dilution that can accompany graph‐based topic communities. However, in the Medium (10–50 words) interval, TGNN edges out our approach: its sentence‐vector graph retains a broader spectrum of semantic nuances, trading off some compression for higher representational entropy that appears beneficial when reviews contain a balanced amount of context.

\begin{table}[t]
\centering
\caption{MSE and MAE Across Varying Review Lengths (Short, Medium, Long, Extra‐long) on Three Datasets. Yelp short‐length results are omitted due to insufficient sample size. The better results are highlighted in \textbf{bold}.}
\label{tab:metrics_no_ratio}

\begingroup
\setlength{\tabcolsep}{4pt}        
\renewcommand{\arraystretch}{0.92} 
\footnotesize                      

\begin{tabular}{@{}cc*{6}{c}@{}}
\Xhline{1pt}
\multirow{3}{*}{\shortstack{\textbf{Review}\\\textbf{Length}}}
  & \multirow{3}{*}{\textbf{Method}}
  & \multicolumn{2}{c}{\textbf{Musical}} 
  & \multicolumn{2}{c}{\textbf{Industrial}} 
  & \multicolumn{2}{c}{\textbf{Yelp}} \\
\cmidrule(lr){3-4} \cmidrule(lr){5-6} \cmidrule(lr){7-8}
& 
& \textbf{MSE} & \textbf{MAE} 
& \textbf{MSE} & \textbf{MAE} 
& \textbf{MSE} & \textbf{MAE} \\
\midrule
\multirow{2}{*}{Short}
  & TGNN         & 0.6912 & 0.4609 & 0.8255 & 0.5245 &   –    &   –     \\
  & TGNN-aspect  & \textbf{0.6485} & \textbf{0.4232} & \textbf{0.7952} & \textbf{0.5098} &   –     &   –    \\
\midrule
\multirow{2}{*}{Medium}
  & TGNN         & 1.1881 & 0.7781 & \textbf{1.3613} & \textbf{0.8515} & \textbf{1.1820} & \textbf{0.7975} \\
  & TGNN-aspect  & \textbf{1.0270} & \textbf{0.7008} & 1.3632 & 0.8598 & 1.2028 & 0.8025 \\
\midrule
\multirow{2}{*}{Long}
  & TGNN         & \textbf{1.1663} & \textbf{0.7514} & \textbf{1.2605} & \textbf{0.7722} & \textbf{1.2423} & 0.8642 \\
  & TGNN-aspect  & 1.2805 & 0.7640 & 1.2640 & 0.7849 & 1.2541 & \textbf{0.8526} \\
\midrule
\multirow{2}{*}{Extra-long}
  & TGNN         & 1.0458 & 0.7694 &  \textbf{1.1954} & 0.7857 & 1.2162 & 0.8628 \\
  & TGNN-aspect  & \textbf{1.0155} & \textbf{0.7246} & 1.1992 & \textbf{0.7792} & \textbf{1.2091} & \textbf{0.8596} \\
\Xhline{1pt}
\end{tabular}
\endgroup
\end{table}

\begin{figure}[t]
    \centering
    \begin{minipage}{0.48\linewidth}
        \centering
        \includegraphics[width=\linewidth]{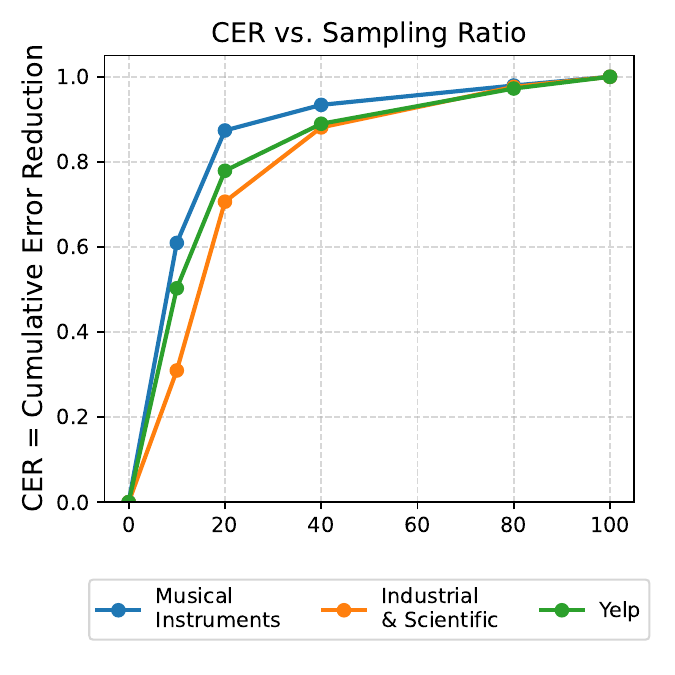}
        \textbf{(a)} CER by sampling ratio on TGNN-aspect.
        \label{fig:ratio_mse}
    \end{minipage}
    \hfill
    \begin{minipage}{0.48\linewidth}
        \centering
        \includegraphics[width=\linewidth]{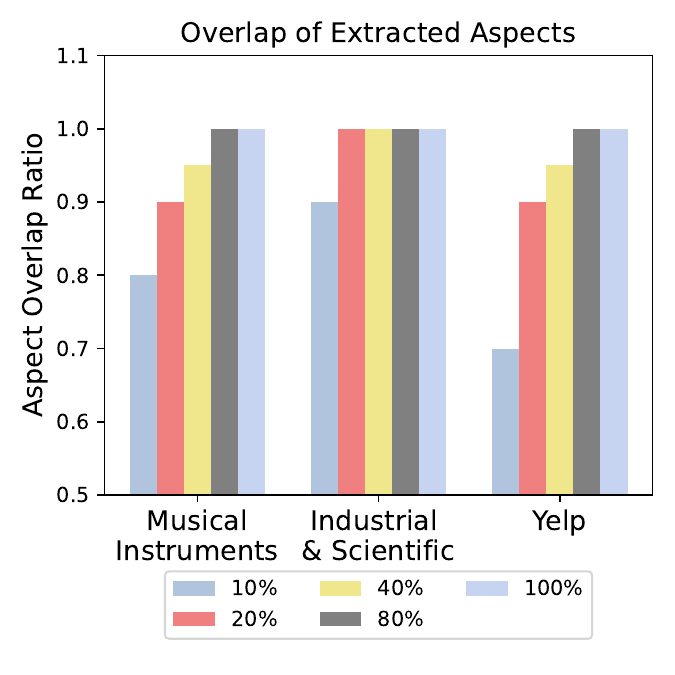}
        \textbf{(b)} Top-20 aspect overlap (\%) with full-data baseline.
        \label{fig:aspect_overlap}
    \end{minipage}
    \caption{Analysis of aspect sampling ratio. (a) CER curves show rapid gains from 10\% to 20\% and diminishing returns thereafter. (b) Overlap between aspects extracted at lower sampling ratios and those extracted from the whole dataset.}
    \label{fig:sampling_analysis}
\end{figure}

\subsubsection{Aspect Sampling Ratio and Dataset Scale}
We investigate how the proportion of sampled reviews and dataset scale influence the quality of extracted aspects and the downstream recommendation performance. For each dataset (Musical Instruments, Industrial and Scientific, and Yelp), we randomly sample a portion of user reviews to perform aspect extraction, varying the sampling ratio among $\{0\%, 10\%, 20\%, 40\%, 80\%, 100\%\}$. The extracted aspects are then used to guide aspect-aware review selection.

To enable a direct comparison of performance gains across datasets, we take the $100\%$ sampling setting as the performance reference and normalize the improvement in MSE relative to the $0\%$ no-aspect-extraction baseline. Specifically, for a sampling ratio $p$, the \emph{cumulative error reduction} (CER) is defined as
\[
\mathrm{CER}(p) = \frac{\mathrm{MSE}(0\%) - \mathrm{MSE}(p)}{\mathrm{MSE}(0\%) - \mathrm{MSE}(100\%)}
\tag{18}
\]
where $\mathrm{MSE}(0\%)$ is the no-aspect baseline and $\mathrm{MSE}(100\%)$ is the full-data condition. CER thus measures the fraction of the total possible MSE reduction (from $0\%$ to $100\%$) that is already achieved at ratio $p$. A value of $\mathrm{CER}(p) \approx 1$ indicates that the performance at ratio $p$ is already close to that of full sampling.

As shown in Fig.~\ref{fig:sampling_analysis}(a), the CER curves rapidly approach $1$ by $20\%$ sampling and flatten thereafter, indicating diminishing returns. In other words, using $20\%$ of reviews for aspect extraction already yields an MSE close to the full-data baseline. This trend is consistent with Fig.~\ref{fig:sampling_analysis}(b), where the Top-20 aspect overlap between $20\%$ extraction and the $100\%$ baseline reach 90\%, suggesting that a small subset of reviews suffices to recover the dominant aspects that drive downstream accuracy.

\subsubsection{Impact of LLM Variants on Semantic Extraction}
To explore the impact of model scale and reasoning depth on semantic extraction, we selected instruction-tuned models from two representative LLM families: the LLaMA 3.x series (3B, 8B, 70B) and the Qwen2.5 series (1.5B, 3B, 7B, 14B, 32B). In addition to these base models, we incorporated DeepSeek-R1-Distill-Llama-8B and
DeepSeek-R1-Distill-Qwen-14B, which are fine-tuned based on samples generated by DeepSeek-R1 to investigate whether long chain-of thought contribute to better aspect-opinion-sentiment extraction. The extracted semantic triples were then injected into three aspect-level recommendation models (EFM, ANR, and TGNN), which are originally designed to leverage topic-level signals derived from raw review text. By replacing their manually constructed or heuristic aspect sources with LLM-derived semantic structures, we systematically examine how different model scales and reasoning strategies affect the performance of aspect-aware recommendation.  

\begin{table*}[t]
\centering
\caption{Impact of LLM Variant and Scale on Aspect-Aware Recommendation (MSE).The best and second-best results are highlighted in bold and underlined.}
\label{tab:performance_mse}
\small
\setlength{\aboverulesep}{0.3ex}   
\setlength{\belowrulesep}{0.25ex}   
\renewcommand{\arraystretch}{0.90} 
\begin{tabular}{ccccc ccc ccc}
\Xhline{1pt}
\multirow{2.5}{*}{\textbf{Model Type}} & \multirow{2.5}{*}{\textbf{Scale}} 
  & \multicolumn{3}{c}{\textbf{EFM-aspect}} 
  & \multicolumn{3}{c}{\textbf{ANR-aspect}} 
  & \multicolumn{3}{c}{\textbf{TGNN-aspect}} \\
\cmidrule(lr){3-5} \cmidrule(lr){6-8} \cmidrule(lr){9-11}
& & Musical & Industrial & Yelp 
  & Musical & Industrial & Yelp 
  & Musical & Industrial & Yelp \\
\midrule
Llama3.x-  
  & 3B   & 1.3953 & 1.6123 & 1.8453 
         & 1.2411 & 1.4932 & 1.3963 
         & 1.1217 & 1.2839 & 1.3421 \\
  & 8B   & \underline{1.3637} & 1.5741 & 1.8275 
         & 1.2348 & \textbf{1.4825} & \underline{1.3958} 
         &  \textbf{1.0960} & 1.2822 & 1.3402  \\
  & 70B  & 1.3648 & 1.5612 & \underline{1.8094} 
         & 1.2334 & 1.4893 & 1.3961 
         & 1.1433 & 1.2828 & 1.3388 \\
\midrule
Qwen2.5-  
  & 1.5B & 1.3729 & \underline{1.5635} & 1.8391 
         & 1.2508 & 1.5064 & 1.3983 
         & 1.1526 & 1.2838 &\textbf{1.3370} \\
  & 3B   & 1.3785 & 1.5676 & 1.8435 
         & \underline{1.2324} & 1.4895 & 1.3969 
         & 1.1496 & 1.2835 & 1.3413 \\
  & 7B   & \textbf{1.3629} & \textbf{1.5494} & \textbf{1.7954} 
         & 1.2423 & 1.4853 & \textbf{1.3956} 
         & 1.1341 & \underline{1.2818} & 1.3395 \\
  & 14B  & 1.3766 & 1.5542 & 1.8205 
         & \textbf{1.2270} & \underline{1.4842} & \underline{1.3958} 
         & \underline{1.1286} & \textbf{1.2612} & \underline{1.3379} \\
  & 32B  & 1.3956 & 1.5924 & 1.8335 
         & 1.2768 & 1.5261 & 1.3977 
         & 1.1475 & 1.2831 & 1.3393 \\
\midrule
DeepSeek‑R1‑Distill-  
  & 8B   & 1.4106  & 1.6632  & 1.8785   
         & 1.2564  & 1.5363  &  1.3967
         & 1.1507  &  1.2865 &  1.3479  \\
  & 14B  & 1.4223  & 1.6057  &  1.8816
         & 1.2612  & 1.5581  &  1.4043 
         & 1.1600  &  1.2942 &  1.3538  \\
\Xhline{1pt}
\end{tabular}
\end{table*}

\begin{figure*}[htbp]  
    \centering
    \includegraphics[width=\linewidth]{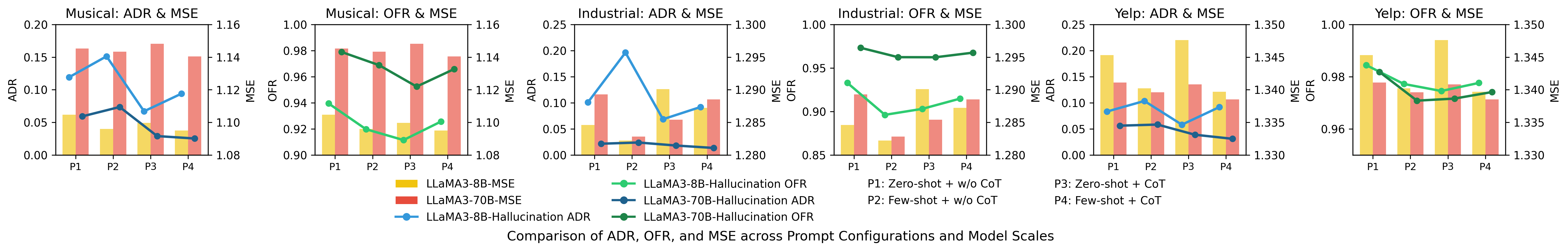}
    \caption{Impact on performance for rating prediction under different prompt design configurations. ($\Delta=2$)}
    \label{fig:prompt_analysis}
\end{figure*}

Table \ref{tab:performance_mse} suggests that increasing the model scale does not guarantee improved performance in aspect-opinion-sentiment extraction. For both LLaMA-3.x and Qwen2.5 families, larger models such as 70B and 32B do not consistently outperform their mid-sized counterparts, suggesting that relatively large models may introduce overfitting to unimportant details or generate overly verbose extractions that hurt downstream recommendation. Conversely, the smallest models (3B and below) exhibit consistently higher errors, indicating insufficient capacity to capture nuanced aspect–opinion–sentiment structures. Finally, when we compare the best-performing base models (LLaMA-3.1-8B and Qwen2.5-14B) against their DeepSeek‑R1‑Distilled versions, the latter uniformly underperform all non-distilled variants. This degradation implies that multi-step CoT optimization, while beneficial for generative reasoning tasks, may inject unnecessary complexity or noise into semantic extraction, thus negatively impacting aspect-aware recommendation.

\subsection{Impact of Hallucination Degree (RQ3)}

\subsubsection{Evaluation Metrics: Aspect Drift Rate (ADR) and Opinion Fidelity Rate (OFR)} 
Table ~\ref{tab:hallucination_metrics} presents \(\mathrm{ADR}\) and \(\mathrm{OFR}\) across the Musical, Industrial, and Yelp datasets for a variety of LLMs and model scales. To assess whether LLM-induced hallucination affects downstream recommendation quality, we correlate ADR/OFR in Table ~\ref{tab:hallucination_metrics} with MSE in Table ~\ref{tab:performance_mse}. Lower hallucination does not uniformly yield better accuracy: mild hallucination often coincides with competitive or improved MSE—likely reflecting useful abstraction/paraphrasing that broadens aspect coverage—whereas extremely low rates are frequently associated with worse MSE, suggesting under-generation or overly conservative outputs caused by overly strict adherence to surface ground-truth, which suppresses the model’s ability to generalize to semantically relevant but lexically divergent expressions and thereby reduces semantic diversity. Overall, the hallucination–performance relationship is non-monotonic: moderate levels can be tolerated or beneficial, but both excessive hallucination and overly strict suppression degrade the extracted semantics and their utility for recommendation.

\begin{table}[t]
\centering
\caption{Hallucination metrics across datasets. The highest hallucination value is highlighted in bold, and the second highest is underlined. ($\Delta=2$)}
\label{tab:hallucination_metrics}

\begingroup
\setlength{\tabcolsep}{4.5pt} 
\renewcommand{\arraystretch}{0.92} 
\setlength{\aboverulesep}{0.3ex}   
\setlength{\belowrulesep}{0.3ex}   
\setlength{\abovetopsep}{0.4ex}    
\setlength{\belowbottomsep}{0.4ex} 
\footnotesize 

\begin{tabular}{@{}lc*{6}{c}@{}} 
\toprule
\multirow{2}{*}{\textbf{Model Type}} & \multirow{2}{*}{\textbf{Scale}} 
  & \multicolumn{2}{c}{\textbf{Musical}} 
  & \multicolumn{2}{c}{\textbf{Industrial}} 
  & \multicolumn{2}{c}{\textbf{Yelp}} \\
\cmidrule(lr){3-4} \cmidrule(lr){5-6} \cmidrule(lr){7-8}
&       & \textbf{ADR} & \textbf{OFR} 
        & \textbf{ADR} & \textbf{OFR} 
        & \textbf{ADR} & \textbf{OFR} \\
\midrule
Llama3.x-  & 3B   & \underline{0.0998} & \textbf{0.7256} & 0.0465  & \textbf{0.7201} & \textbf{0.0956} & 0.9449\\
           & 8B   & \textbf{0.1153} & 0.9121 & \textbf{0.1606}  & 0.8770 & \underline{0.0804} & 0.9892\\
           & 70B  & 0.0674 & 0.9852 & 0.0242  & 0.9828 & 0.0577 & 0.9896\\
\midrule
Qwen2.5-   & 1.5B & 0.0865 & 0.9007 & \underline{0.0579}  & 0.9284 & 0.0585 &  \underline{0.9276}\\
           & 3B   & 0.0294 & \underline{0.7273} & 0.0421  & \underline{0.7727} & 0.0523 & 0.9450\\
           & 7B   & 0.0775 & 0.9199 & 0.0335  & 0.9161 & 0.0419 & 0.9515\\
           & 14B  & 0.0322 & 0.9153 & 0.0263  &  0.9361 & 0.0398 & 0.9555\\
           & 32B  & 0.0073 & 0.9667 & 0.0145  & 0.9782 & 0.0042 & 0.9904\\
\midrule
Distill-   & 8B   & 0.0541 & 0.8480 & 0.0370 & 0.8485 & 0.0192 & \textbf{0.8743}\\
           & 14B  & 0.0124 & 0.9289 & 0.0117 & 0.9366 &  0.0156 & 0.9305\\
\bottomrule
\end{tabular}
\endgroup
\vspace{-6pt}
\end{table}

\subsubsection{Prompt Design for Aspect Extraction.}

To systematically examine the impact of prompt design on hallucination severity and downstream recommendation performance, we construct a controlled experiment varying both the number of exemplars (zero-/few-shot) and reasoning depth (with/without Chain-of-Thought, CoT). Specifically, we apply four prompt configurations—Zero-shot w/o CoT, Few-shot w/o CoT, Zero-shot + CoT, and Few-shot + CoT—across two representative LLM variants: LLaMA3.1-8B-Instruct and LLaMA3.3-70B-Instruct. Each model is prompted to extract aspect-opinion pairs from review text under a fixed aspect schema, enabling direct measurement of both extraction fidelity and generalization behavior.

As shown in Fig.~\ref{fig:prompt_analysis}, all prompt configurations display analogous behaviors across the three datasets. The few-shot without CoT setting consistently yields the highest ADR, highlighting its tendency to select aspects outside the constrained schema. By contrast, adding CoT reduces ADR; the lowest ADR is obtained by zero-shot+CoT on 8B and by few-shot+CoT on 70B, indicating stronger aspect grounding. This gain is accompanied by reduced OFR: CoT often increases opinion-level speculation with the decrease most evident on Musical and Industrial, and still notable on Yelp-70B. Model capacity moderates this trade-off: under few-shot+CoT, the 70B variant attains the lowest ADR while keeping OFR high, yielding the most balanced configuration overall. Conversely, few-shot without CoT not only maximizes aspect drift but also often produces lower OFR—especially on Industrial-8B and Yelp-70B—underscoring the need to balance semantic coverage, factual grounding, and generation fidelity given both model scale and task requirements.

\section{Conclusions and future work}
In this work, we proposed the Hyper-Adaptive Dual-Stage Semantic Framework that first condenses raw user reviews into a concise, high‐quality aspect vocabulary, then uses this vocabulary to guide LLMs in extracting accurate aspect–opinion triples. This approach consistently improves both rating prediction across varied review densities and lengths. We further introduced two interpretable hallucination metrics—Aspect Drift Rate (ADR) and Opinion Fidelity Rate (OFR)—and showed how varying abstraction levels affect recommendation quality in a non‐linear manner. Our systematic study of prompt configurations, covering zero‐/few‐shot and chain‐of‐thought settings, provides actionable guidelines for balancing semantic fidelity and expressive coverage. Together, these contributions enhance the robustness and explainability of review‐aware recommendation systems, while opening avenues for future work on adaptive prompt tuning, multimodal feedback integration, and principled hallucination control in practical deployments.

\section{Acknowledgments}
This work is funded by Jiangsu Provincial Young Science and Technology Talent Support Program(Grants No. JSTJ-2025-944), Science and Technology Major Special Program of Jiangsu (Grants No. BG2024028)
\clearpage   
\appendix
\appendix
\section{Additional Details}
\subsection{Baseline Methods}\label{app:baselines}
For completeness, we briefly describe the baseline methods evaluated in this work:
\begin{itemize}[leftmargin=*, align=left, itemsep=0pt, topsep=0pt, parsep=0pt]
    \item \textbf{PMF}~\cite{mnih2007probabilistic} aims to solve for the probability distribution parameters
of users and items by maximizing the posterior probability of the
training data.
    \item \textbf{EFM}~\cite{zhang2014explicit} extracts feature–opinion aspects via phrase-level sentiment analysis to provide explainable reasons for recommendations.
    \item \textbf{ANR}~\cite{10.1145/3269206.3271810} performs aspect-based representation learning for both
users and items via an attention-based component.
    \item \textbf{DeepCoNN}~\cite{Zheng2017JointDM} jointly models user behavior and product attributes from textual reviews.
    \item \textbf{NARRE}~\cite{Chen2018NeuralAR} estimates the usefulness of each review with attention and uses these weights for prediction.
    \item \textbf{RGCL}~\cite{shuai2022review} constructs a review-aware user-item graph with feature-enhanced edges and applying contrastive learning tasks to better
model user-item interactions.
    \item \textbf{TGNN}~\cite{Shuai2023TopicenhancedGN} introduces topic-enhanced and review-enhanced rating graphs to analyze explicit and implicit review information.
    \item \textbf{GPT-4o}~\cite{liu2023chatgpt} uses GPT-4o to directly predict ratings based on the user’s interaction (review, rating, etc.). We adopted the same few-shot prompting template as~\cite{liu2023chatgpt}.
    \item \textbf{Rec-SAVER}~\cite{tsai2024leveragingllmreasoningenhances} augments the LLM’s input with
the target item’s metadata and the user’s past interaction history,
and then prompts the model to articulate its reasoning steps
before producing its prediction.
    \item \textbf{EXP3RT}~\cite{kim2024driven} extracts preference from reviews to build user and
item profiles, and employs chain-of-thought reasoning to generate accurate and explainable rating predictions.
\end{itemize}

\subsection{Implementation Details}
For semantic aspect extraction, we employ the LLaMA3.3-70B-Instruct model\footnote{\url{https://www.llama.com/docs/model-cards-and-prompt-formats/llama3_3/}} to identify salient aspects from user reviews. To ensure robustness and diversity, we randomly sample 20\% of the entire dataset and repeat the extraction process five times. The top frequent aspects are then selected to guide the downstream review filtering process. Based on these extracted aspects, we utilize LLaMA3.1-8B-Instruct model\footnote{\url{https://huggingface.co/meta-llama/Llama-3.1-8B-Instruct}} to perform fine-grained semantic-aware review extraction. 

For all LLM-based recommendation models, we fine-tune LLaMA3-8B model\footnote{\url{https://www.llama.com/docs/model-cards-and-prompt-formats/meta-llama-3/}} using the QLoRA technique. For other baseline methods, we adopt the official implementations from previous work. For fair comparison, we conduct each experiment five times using different random seeds, and report the average values. 

\section{Case study}
\begin{figure}[t]
    \centering
    \begin{minipage}{0.9\linewidth}
        \centering
        \includegraphics[width=\linewidth]{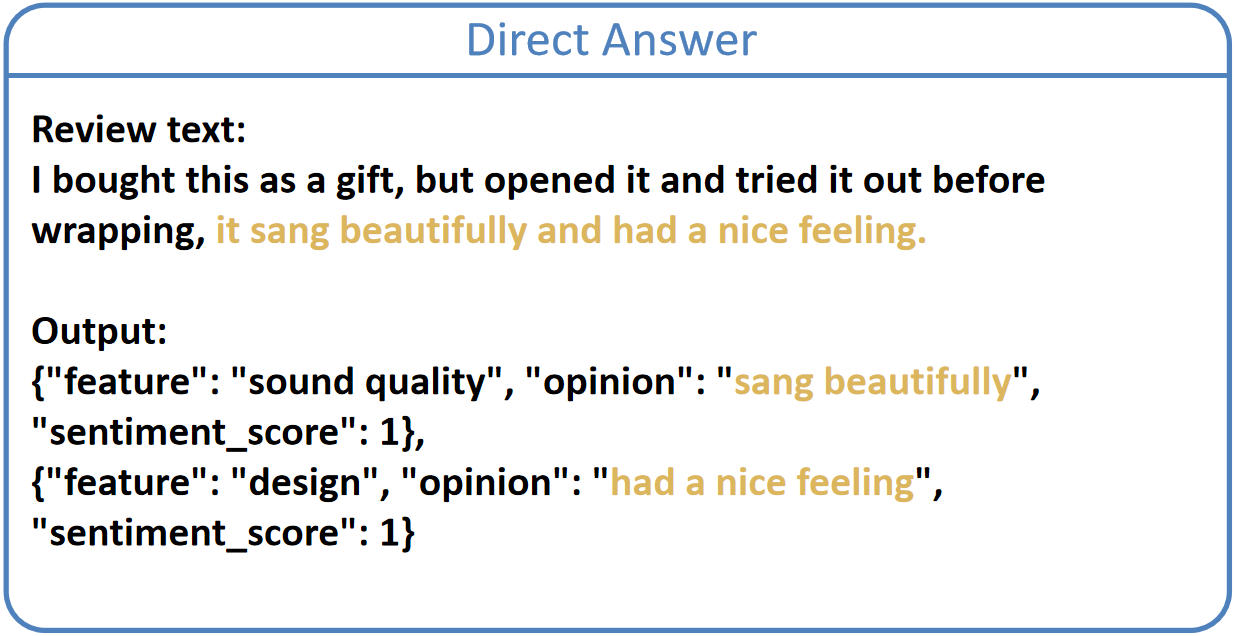}
        \caption*{\textbf{(a)} Direct Answer (Generate by LlaMA3.1-8B-Instrct from Musical Instrument.}
    \end{minipage}
    \vspace{0.6em}

    \begin{minipage}{0.9\linewidth}
        \centering
        \includegraphics[width=\linewidth]{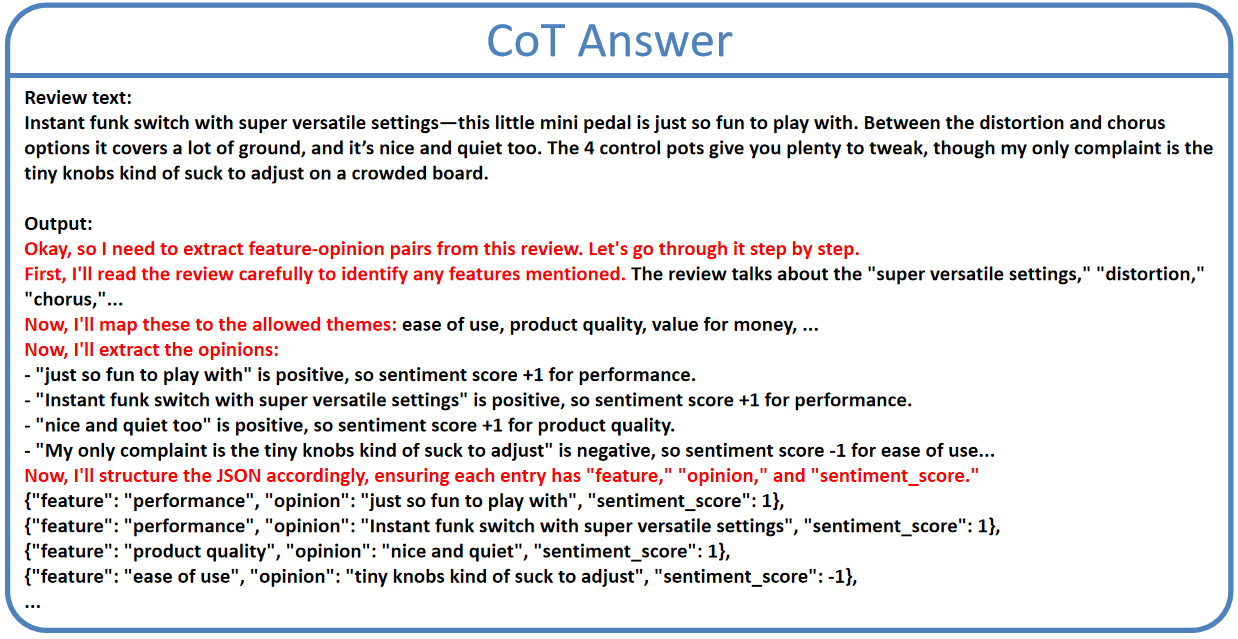}
        \caption*{\textbf{(b)} CoT Answer (Generate by DeepSeek-R1-Distill-LlaMA-8B from Industrial and Scientific.}
        \label{fig:aspect_overlap}
    \end{minipage}
    \vspace{0.6em}

    \begin{minipage}{0.9\linewidth}
        \centering
        \includegraphics[width=\linewidth]{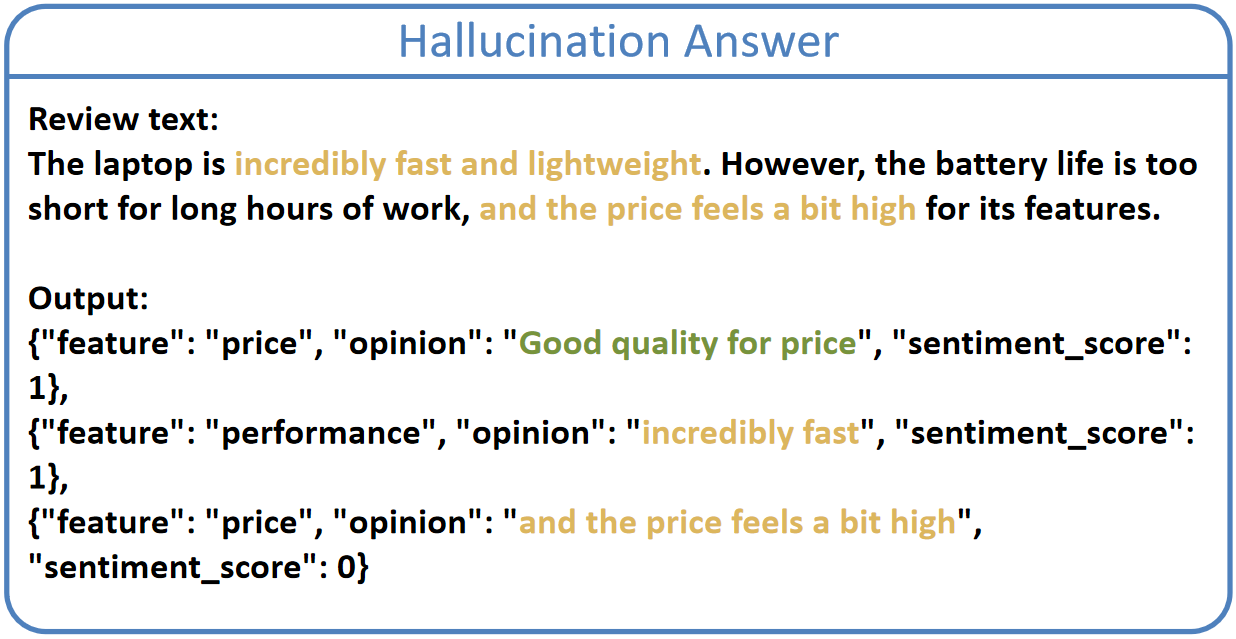} 
        \caption*{\textbf{(c)} Hallucination Answer (Harmfully generate by LlaMA3.2-3B-Instrct from  Industrial and Scientific.} 
    \end{minipage}

    \caption{Case study of different prompting strategies for aspect-opinion extraction. (a) Direct Answer shows faithful extraction without explicit reasoning. (b) CoT Answer demonstrates step-by-step reasoning to ensure alignment with predefined aspects. (c) Hallucination Answer contains ungrounded opinions introduced by flawed few-shot exemplars.}
    \label{fig:case_study}
\end{figure}

 Figure~\ref{fig:case_study} compares three representative outputs under different prompting strategies: Hallucination Answer, Direct Answer, and CoT Answer. The green highlights indicate hallucinated opinions not grounded in the input text, typically introduced by flawed few-shot exemplars. In contrast, yellow highlights show faithfully extracted opinions aligned with the original review. The red text in the CoT Answer illustrates a step-by-step reasoning process that ensures alignment with predefined aspects and improves transparency. Compared to direct or few-shot prompting, CoT prompting effectively mitigates harmful hallucinations and enhances the reliability of aspect-opinion extraction.

\clearpage


\bibliographystyle{ACM-Reference-Format}
\bibliography{sample-base}
\clearpage
\section*{Ethical Considerations}
In this section, we explore the ethical implication of Hyper‑Adaptive Dual‑Stage Semantic Framework (HADSF), which induces a corpus‑level aspect vocabulary and performs controlled extraction of aspect–opinion triples from user reviews while diagnosing hallucination via Aspect Drift Rate (ADR) and Opinion Fidelity Rate (OFR). We discuss considerations for implementation and deployment at scale.

\textbf{Data Privacy and Consent.}
HADSF operates on large volumes of user‑generated reviews to construct structured, explanatory signals for recommendation. Deployments should follow strict data‑minimization principles, confining inputs to publicly available review text and pseudonymous IDs, and must not attempt re‑identification. Strong safeguards—access control, encryption in transit/at rest, and periodic security audits—are required. When producing auxiliary annotations or synthetic labels, practitioners should document data lineage, honor dataset terms and applicable regulations (e.g., GDPR/CCPA), and prefer aggregation, pseudonymization, and (where feasible) differential privacy to mitigate leakage risks. Any released artifacts (code, prompts, metric implementations) should exclude raw personal data and preserve the original datasets’ usage constraints.

\textbf{Reliability and Transparency.}
LLM‑based extraction can introduce fabrication or bias that propagates to downstream recommenders. HADSF addresses this by constraining extraction with a learned aspect vocabulary and auditing outputs using ADR (to detect aspect drift) and OFR (to verify grounding in source spans). In practice, teams should set thresholding/early‑halt policies on these metrics, perform regular bias audits across user/item strata, and report not only rating‑error gains but also stability under varied review lengths and domains. Explanations should link recommended items to specific review spans and aspect terms to support user recourse and oversight. Given our analysis across models from 1.5B to 70B parameters, practitioners should also consider cost–quality trade‑offs: prefer the smallest model that meets pre‑specified reliability thresholds to reduce operational risk and environmental footprint while maintaining transparency about compute budgets and evaluation protocols.


\end{document}